\documentclass[10pt,twocolumn,letterpaper]{article}
\usepackage[margin=0.75in,columnsep=0.25in]{geometry}
\usepackage{times}
\usepackage[T1]{fontenc}
\usepackage[utf8]{inputenc}
\usepackage{amsmath,amssymb}
\usepackage{graphicx}
\usepackage{booktabs}
\usepackage{hyperref}
\usepackage{caption}
\usepackage[numbers,round]{natbib}
\usepackage{authblk}

\title{\bf The Density of Thought: Predicting and Steering\\
Reasoning in Large Language Models}
\author[1]{Tanush Kalhan\thanks{Corresponding Author: Tanush.Kalhan.ug24@plaksha.edu.in}}
\author[2]{Anupam Chattopadhyay}
\author[2]{Sriram Ranga}
\author[3,4]{Amritendu Mukherjee}

\affil[1]{Plaksha University}
\affil[2]{Nanyang Technological University}
\affil[3]{Interview Street Technologies Pvt. Ltd. (HackerRank)}
\affil[4]{Indian Statistical Institute}

\date{}

\begin{document}
\maketitle

\begin{abstract}
The Uniform Information Density hypothesis holds that effective
communication distributes information evenly across an utterance.
We investigate whether variation in information density across
chain-of-thought (CoT) reasoning steps predicts answer correctness
and can serve as a controllable inference-time target. We prompt
frozen pretrained models to solve problems via CoT and measure the
density of each reasoning step with two independent metrics: LP
variance (how unevenly information is distributed across steps,
computed as the variance in mean per-token negative log-likelihood)
and the Caveman compression ratio. We then test whether density
correlates with, and can be steered to improve, answer correctness.
Experiments across six model scales, two training paradigms, and 60
GSM8K problems yield four main findings. First, lower LP variance
predicts correctness in supervised fine-tuned (SFT) models, but this
relationship weakens with scale and reverses in reinforcement-learning
(RL) distilled models. Training paradigm therefore appears more
influential than model size in determining density behaviour. Second,
synthetic experiments reveal an intrinsic complexity preference:
models trained on uniformly distributed data learn
complexity-favouring distributions 2.0--2.6 times more accurately than
simplicity-favouring ones. Third, direct inference-time density
steering reduces accuracy from 60.0\% to 8.3\%, indicating that
natural density patterns are functionally important to reasoning.
Fourth, pre-generation nudging combining hint injection with
contrastive KV-cache steering improves accuracy by 13.3 percentage
points, consistent across both metrics. Finally, we introduce
Entropy-Gated Density Routing (EGDR), a self-supervised method using
token-level Shannon entropy to gate proportional KV-cache adjustments
during generation, improving accuracy by 10.0 percentage points
without any external guidance.\\
Code is available at \url{https://github.com/tanushhh21/density_of_thought}.

\end{abstract}

\section{Introduction, Motivation, and Contributions}

\paragraph{Context and problem statement.}
Chain-of-thought (CoT) prompting \citep{wei2022cot} has become the
dominant strategy for eliciting multi-step reasoning from large
language models (LLMs), improving performance on arithmetic, symbolic,
and commonsense tasks by decomposing complex problems into explicit
intermediate steps. Despite this empirical success, we lack a
principled understanding of \emph{why} a given reasoning trace
succeeds or fails. Two traces produced for the same problem can be
nearly identical in length, vocabulary, and surface style, yet arrive
at opposite answers. This suggests that correctness is not determined
solely by \emph{what} a model writes, but by \emph{how} it organises
the flow of information across its reasoning steps. The central
problem we address is therefore the following: is there a measurable,
structural property of a CoT trace that reveals whether the underlying
reasoning is sound, and if so, can that property be controlled at
inference time to make reasoning more reliable?

\paragraph{Information density in reasoning.}
We propose that the relevant property is \emph{information density}:
the degree to which each reasoning step contributes new,
hard-to-predict content rather than predictable filler. Reasoning
steps are not informationally uniform. A step that performs a genuine
computation, such as evaluating an arithmetic expression or applying a
rule, is difficult for the model to predict and therefore carries high
information content. A step that merely restates the problem or
provides connective phrasing is easy to predict and carries little.
Because CoT reasoning unfolds autoregressively, the density of each
step is shaped by, and in turn shapes, the internal state the model
carries forward through generation. This coupling makes density a
natural candidate for a correctness signal: the way information is
distributed across steps may reflect whether the model is genuinely
reasoning or merely producing plausible-looking text.

\paragraph{Motivation from Uniform Information Density.}
This intuition has a well-established theoretical grounding in
psycholinguistics. The Uniform Information Density (UID) hypothesis
\citep{levy2007uid} holds that efficient communicators distribute
information as evenly as possible across an utterance, avoiding both
dense spikes that overload the listener and sparse stretches that
waste communicative capacity. UID has been validated across languages
and discourse contexts \citep{tsipidi2024surprise,aceves2024denser},
and it offers a concrete, testable prediction when transferred to
machine reasoning: if reasoning behaves like communication, then
correct traces should exhibit smoother, more uniform information
density than incorrect ones. We take this prediction as our starting
point and ask whether it holds for LLM reasoning, how it interacts
with a model's scale and training, and whether it can be turned from a
passive descriptor into an active lever for improving accuracy.

\paragraph{Findings and framing.}
Our answer is nuanced. Information density \emph{does} signal
reasoning quality, but the direction of the signal is not universal:
it depends on the paradigm under which a model was trained rather than
on its size. More importantly, we find that density is not a
superficial stylistic feature that can be freely rewritten. It is
\emph{functionally load-bearing}, encoding the model's accumulated
computational context within the key-value (KV) cache, so that
overriding it mid-generation destroys reasoning coherence rather than
improving it. This distinction, between density as a \emph{signal} of
reasoning quality and density as a \emph{tool} for influencing it,
organises the paper. It also yields a design principle for
inference-time control: effective intervention must work \emph{with} a
model's natural density organisation rather than against it.

\paragraph{Contributions.}
Mirroring the structure of our findings, this paper makes the
following contributions:

\begin{itemize}
\item \textbf{Density as a regime-dependent signal.} We present the
first systematic study of information density as a correctness signal
in CoT reasoning across six model scales and two training paradigms,
establishing that lower LP variance predicts correctness under
supervised fine-tuning but reverses under reinforcement-learning
distillation, so training paradigm is more influential than model
size.

\item \textbf{An intrinsic complexity preference.} Through controlled
synthetic experiments following \citet{ranga2026density}, we show that
LLMs possess an intrinsic structural bias toward syntactically
complex, high-density text, a 2.0--2.6$\times$ KL asymmetry consistent
across all model sizes, confirming that this bias is a property of the
learning process rather than of training data.

\item \textbf{Density is functionally load-bearing.} We demonstrate
that na\"{i}ve density steering fails catastrophically ($-51.7$pp) and
derive a precise design principle, \emph{preserve step-level
generation integrity}, that explains the failure and guides all
subsequent methods.

\item \textbf{Density as an effective tool.} We introduce a principled
nudging pipeline (NuRL-style hint injection with contrastive KV cache
steering) that improves accuracy by $+13.3$pp on GSM8K, with density
footprints confirmed by two independent measurements, and EGDR, a
self-supervised method that uses per-token Shannon entropy to gate
proportional KV cache correction during generation, improving accuracy
by $+10.0$pp without any external input ($p{=}0.019$).
\end{itemize}

\section{Related Work}

\paragraph{Uniform Information Density in language and cognition.}
The UID hypothesis \citep{levy2007uid} predicts that speakers minimise
peak processing effort by spreading information evenly across
utterances. \citet{tsipidi2024surprise} show that discourse structure
independently modulates density: speakers are informationally denser
at discourse peaks and sparser at transitions.
\citet{aceves2024denser} provide cross-linguistic evidence that denser
languages communicate faster but with narrower conceptual focus. This
trade-off is relevant to CoT reasoning, where denser steps commit to
narrower but deeper computational paths.

\paragraph{Information density in LLM reasoning.}
\citet{gwak2025revisiting} revisit the UID hypothesis in LLM
reasoning, finding that selecting traces with more uniform step-level
density improves accuracy by 10--32\% on AIME2025. Their approach
selects among \emph{completed} traces; EGDR shapes density
\emph{during} generation. \citet{zhao2025denser} identify a
dual-density structure: computational steps should be dense while
explanatory steps should be sparse. This motivates our finding that
na\"{i}ve global density steering fails: because different steps
require different densities, any intervention that homogenises them
destroys the structure that makes reasoning coherent.

\paragraph{LLM density modelling.}
\citet{ranga2026density} study distributional learning in LLMs using a
synthetic number language, finding that models learn syntactically
cued distributions but fail on semantic ones. Our synthetic experiment
(detailed in the Signal section) replicates their uniform-corpus KL
result ($0.0038 \pm 0.0010$) and extends their framework to
complexity-asymmetric distributions. \citet{ranga2025plagiarism}
conjecture that AI-generated text will eventually saturate
high-probability sentence space (Plagiarism Singularity). Our
cross-generational evidence shows this space is contracting faster
than their GPT-2-based estimate assumes.

\paragraph{CoT improvement methods.}
Self-consistency \citep{wang2023selfconsistency} aggregates multiple
sampled traces via majority voting; reranking methods select the
highest-reward completed trace. Both operate over completed traces and
incur sampling overhead. Our approach is orthogonal: we steer a single
greedy generation through density-aware KV correction, adding
$\approx 0.3\%$ per-token overhead.

\paragraph{Inference-time steering and prompt compression.}
KV cache steering controls style, safety, and topic via fixed global
vectors \citep{turner2023activation,hernandez2023inspecting}.
\citet{li2025entropy} use per-token entropy to gate generation
branching. EGDR differs in that it applies a continuous, proportional,
in-place KV correction during generation, rather than branching,
post-hoc selection, or a fixed global vector. We repurpose Caveman
Compression \citep{peltomaki2025caveman} as a density
\emph{measurement} tool for generated reasoning steps rather than for
input compression, providing a signal independent of model
probabilities.

\section{Preliminaries and Density Measurement Framework}

\subsection{Information Density in Reasoning Traces}

A CoT reasoning trace consists of $K$ steps $s_1, \dots, s_K$, each a
sequence of tokens generated autoregressively. Not all steps are
informationally equal: a step computing an arithmetic result
contributes novel, hard-to-predict content, while a transitional
phrase is nearly deterministic given context. This variation in
\emph{information density} is the central object of our study.

The UID hypothesis provides the grounding for this study: it predicts
that effective communicators distribute information evenly, avoiding
both dense spikes (which cause processing errors) and sparse passages
(which waste communicative capacity). Applied to CoT, we ask whether
the same principle holds: are correct reasoning chains more uniformly
dense, and can we steer density to improve correctness?

\subsection{Density Metrics}

We measure density via two complementary and independent signals.

\paragraph{Log-probability (LP) variance (model-internal).}
The LP score of step $s_k$ is the mean per-token negative
log-likelihood under the generating model:
\begin{equation}
\mathrm{LP}(s_k) = \frac{1}{|s_k|} \sum_{t \in s_k}
-\log P(t \mid s_{<t}, \text{prompt}).
\end{equation}
Higher $\mathrm{LP}(s_k)$ indicates greater information density: the
model found the step harder to predict. We compute step-level LP
variance after min--max normalisation:
$v = \mathrm{Var}(\widehat{\mathrm{LP}}(s_1), \dots,
\widehat{\mathrm{LP}}(s_K))$, where
$\widehat{\mathrm{LP}}(s_k) = (\mathrm{LP}(s_k) - \min_j
\mathrm{LP}(s_j)) / (\max_j \mathrm{LP}(s_j) - \min_j
\mathrm{LP}(s_j) + \epsilon)$. Z-score normalisation is excluded: it
forces variance to 1.0 for every chain by definition, eliminating all
between-chain signal.

\paragraph{Caveman compression ratio (linguistically grounded).}
The Caveman ratio $\rho(s_k)$ \citep{peltomaki2025caveman} is the
fraction of tokens removed by a rule-based SpaCy pipeline that strips
predictable grammatical tokens (determiners, auxiliary verbs,
coordinating conjunctions, intensifiers). Lower $\rho$ indicates fewer
predictable tokens and hence higher density. This signal is computed
entirely from the surface form of the generated text, with no
reference to model probabilities, and is therefore fully independent
of LP variance.

\paragraph{Why two signals.}
LP variance is model-internal; Caveman ratio is linguistically
grounded. They share no computational basis. When both agree in
direction across conditions, the density effect is real rather than a
measurement artefact.

\begin{figure*}[t]
\centering
\includegraphics[width=\textwidth]{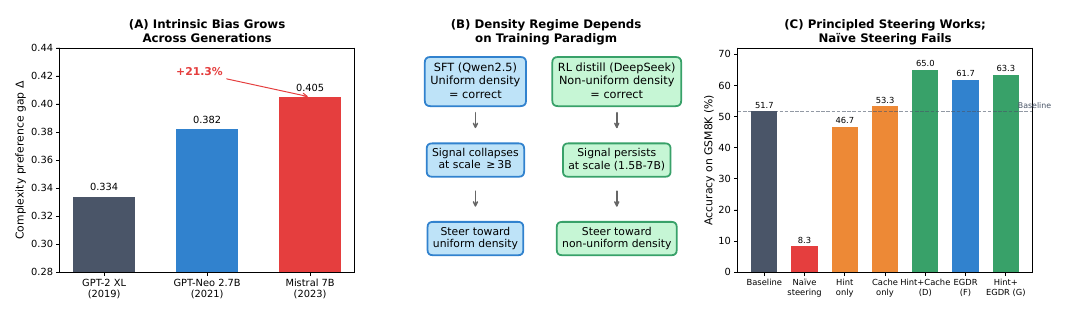}
\caption{Overview of the paper's argument. (A) The complexity
preference gap $\Delta$ grows 21.3\% from GPT-2 XL (2019) to Mistral
7B (2023), driven by training quality rather than scale alone. (B) The
density--correctness relationship depends on training paradigm: SFT
models show uniform density in correct solutions (consistent with
UID); RL-distilled models show structured non-uniformity. This
determines the beneficial direction of any density-based intervention.
(C) Na\"{i}ve density steering collapses accuracy by 51.7\,pp;
principled pre-generation nudging (D: Hint+Cache) achieves
$+13.3$\,pp; EGDR achieves $+10.0$\,pp without any external hint.}
\label{fig:overview}
\end{figure*}

\section{Information Density as a Signal}

We first establish why information density varies systematically
across model families, and whether this variation carries diagnostic
information about reasoning quality.

\subsection{LLMs Exhibit a Growing Complexity Preference}

Before examining density in reasoning traces, we document a
foundational empirical phenomenon: language models exhibit a growing
preference for syntactically complex, high-density text that has
strengthened markedly across model generations. This preference is not
trivially explained by training data: a controlled synthetic
experiment confirms the bias is intrinsic to the learning process
itself.

\paragraph{Cross-generational study.}
We construct a corpus of 3,204 Wikipedia sentences filtered for length
(8--40 tokens) and grammatical completeness. Syntactic complexity is
measured via SpaCy dependency tree depth; sentences are binned into
low ($\leq 5$), mid (5--7), and high ($> 7$) tiers. For ten pretrained
models spanning GPT-2 Small (124M, 2019) through Mistral 7B (2023), we
compute the complexity preference gap:
$\Delta = \mathrm{LP}(\textsc{high}) - \mathrm{LP}(\textsc{low})$.
Three model families disentangle scale from generational effects:
GPT-2 (same training data, varying scale), Pythia (same architecture
and data, controlled scale), and a cross-generational comparison
(different architectures, training regimes, and years).

Table~\ref{tab:crossgen} shows that the cross-generational effect
($+21.3\%$) substantially exceeds within-family scaling ($+13.2\%$
over $40\times$ parameters in the controlled Pythia family). The GPT-2
XL to GPT-Neo transition is notably weak ($p{=}0.187$, \textsc{low}
tier), localising the major shift to 2021--2023. The probability ratio
$e^{\Delta}$, which quantifies how much more likely a high-complexity
sentence is than a low-complexity one, grew from 1.40 (2019) to 1.50
(2023).

\begin{table}[t]
\centering\small
\caption{Complexity preference gap $\Delta$ across model families. The
cross-generational effect ($+21.3\%$) substantially exceeds the
controlled within-family scaling effect ($+13.2\%$ over $40\times$
parameters in Pythia), implicating training methodology, not scale, as
the primary driver. Full per-model results in
Table~\ref{tab:crossgen-full}.}
\label{tab:crossgen}
\begin{tabular}{lccc}
\toprule
Comparison & Base $\Delta$ & Max $\Delta$ & Change \\
\midrule
GPT-2 (124M$\rightarrow$1.5B)      & 0.3190 & 0.3378 & $+4.8\%$ \\
Pythia (70M$\rightarrow$2.8B, ctrl.) & 0.2935 & 0.3323 & $+13.2\%$ \\
Cross-gen. (2019$\rightarrow$2023) & 0.3343 & 0.4054 & $+21.3\%^{***}$ \\
\bottomrule
\end{tabular}

\vspace{2pt}
{\scriptsize $^{***}$ KS: \textsc{low} $p{=}1.3\times10^{-47}$,
\textsc{mid} $p{=}1.6\times10^{-47}$,
\textsc{high} $p{=}2.8\times10^{-31}$.}
\end{table}

\paragraph{Intrinsic bias: synthetic evidence.}
Does the complexity preference arise from training data or from the
learning process itself? Following \citet{ranga2026density}, we train
four GPT-2-like models (2M, 6M, 25M, 50M parameters) from scratch on a
synthetic corpus with a fixed 67-word vocabulary and sentences at
dependency depths 2--8. We use three training distributions: Uniform
($P(d){=}1/7$), Linear ($P(8){=}4{\times}P(2)$,
complexity-favouring), and Inverse ($P(2){=}4{\times}P(8)$,
simplicity-favouring). Only the distribution over depth levels
changes; vocabulary and grammar are identical.

\begin{table}[t]
\centering\small
\caption{Synthetic experiment results (averaged across 2M--50M
models). The Uniform KL (0.0038) replicates
\citet{ranga2026density}. The KL asymmetry ratio (Inverse/Linear
$= 2.4\times$) is consistent across all individual model sizes
(range: 2.0--2.6$\times$).}
\label{tab:synthetic}
\begin{tabular}{lcccc}
\toprule
Corpus & KL (avg) & Bias & Spear.\ $\rho$ & KL ratio \\
\midrule
Uniform & 0.0038 & $+0.094$ & $0.60^{*}$     & --            \\
Linear  & 0.0409 & $-0.50$  & $0.964^{***}$  & $1.0\times$   \\
Inverse & 0.0967 & $+0.77$  & 0.21 n.s.      & $2.4\times$   \\
\bottomrule
\end{tabular}

\vspace{2pt}
{\scriptsize $^{*}p<0.05$; $^{***}p<0.001$. Bias: signed deviation of
learned $P(d)$ from ground truth.}
\end{table}

Table~\ref{tab:synthetic} reveals two principal findings. Under
Uniform training, all models assign higher probability to complex
sentences despite equal training representation (positive bias,
$+0.077$ to $+0.104$), establishing the bias is intrinsic. The KL
asymmetry ratio $\mathrm{KL}(\text{Inverse})/\mathrm{KL}(\text{Linear})$,
which ranges from 2.0 to $2.6\times$ across all model sizes, shows
that models learn complexity-favouring distributions more than twice
as accurately as simplicity-favouring ones under otherwise identical
conditions. Under Inverse training, models simultaneously
under-represent simple sentences and over-represent complex ones,
actively resisting the intended distribution at both ends. This
structural bias means that every downstream task, including CoT
reasoning, begins with a pre-existing pull toward dense text. Whether
this pull manifests as a measurable correctness signal in reasoning is
the question we address next.

\subsection{Density Variance Predicts Correctness, but Paradigm
Determines Direction}

\paragraph{Setup.}
We evaluate on GSM8K \citep{cobbe2021gsm8k} ($N{=}60$, greedy
decoding) using two model families sharing the Qwen base architecture:
Qwen2.5-Instruct (0.5B, 1.5B, 3B, 7B; standard SFT) and DeepSeek
R1-Distill-Qwen (1.5B, 7B; RL distillation). For each model and
problem, we compute LP variance and measure the LP gap
$\Delta_{\mathrm{LP}} = \mathrm{Var}(C) - \mathrm{Var}(W)$, where
$\mathrm{Var}(C)$ and $\mathrm{Var}(W)$ are mean LP variances over
correct and incorrect solutions respectively.

\begin{table}[t]
\centering\small
\caption{Density--correctness relationship across model families. LP
gap $= \mathrm{Var}(C) - \mathrm{Var}(W)$. Architecture, vocabulary,
tokeniser, and inference procedure are identical between SFT and RL
families at each scale; the reversal is attributable to training
paradigm.}
\label{tab:paradigm}
\begin{tabular}{llccl}
\toprule
Model & Paradigm & Acc. & $\Delta_{\mathrm{LP}}$ & Regime \\
\midrule
Qwen2.5-0.5B & SFT & 26.7\% & $-0.029$ & Reversed  \\
Qwen2.5-1.5B & SFT & 51.7\% & $+0.020$ & UID       \\
Qwen2.5-3B   & SFT & 60.0\% & $-0.003$ & Collapsed \\
Qwen2.5-7B   & SFT & 63.3\% & $-0.001$ & Collapsed \\
\midrule
R1-Distill-1.5B & RL & 66.7\% & $+0.013$ & Dual-density \\
R1-Distill-7B   & RL & 76.7\% & $+0.014$ & Dual-density \\
\bottomrule
\end{tabular}
\end{table}

\paragraph{Standard SFT: the UID regime.}
At Qwen2.5-1.5B, $\Delta_{\mathrm{LP}} = +0.020$: correct solutions
have lower LP variance, consistent with UID. However, this signal
collapses monotonically with scale: by 3B
($\Delta_{\mathrm{LP}}{=}-0.003$) and 7B
($\Delta_{\mathrm{LP}}{=}-0.001$), LP variance is virtually identical
for correct and incorrect solutions. Larger SFT models produce
uniformly dense text regardless of correctness, erasing the density
signature of reasoning quality.

\paragraph{RL distillation: the dual-density regime.}
DeepSeek R1-Distill-Qwen reverses the relationship. At 1.5B,
$\Delta_{\mathrm{LP}} = +0.013$: correct solutions have higher LP
variance, with dense computational peaks and sparse explanatory
troughs, a pattern consistent with the dual-density structure of
\citet{zhao2025denser}. Critically, this signal persists across scale
($\Delta_{\mathrm{LP}} = +0.014$ at 7B), unlike the SFT signal. Both
RL-distilled models also outperform their SFT counterparts at matched
scale (Table~\ref{tab:paradigm}).

\paragraph{Implication.}
Training paradigm, not parameter count, determines whether UID or
dual-density describes a model's correct-reasoning profile. Any
intervention designed to improve reasoning via density must therefore
be calibrated to the model's paradigm, since the direction of
beneficial density change is not universal. Before designing any
intervention, however, we must understand why the most obvious
approach fails, and what that failure reveals about the functional
nature of density in reasoning.

\section{Information Density as a Tool}

\subsection{Na\"{i}ve Steering Fails, and Reveals Why}

The density--correctness correlation at Qwen2.5-1.5B suggests an
obvious intervention: steering density toward the profile of correct
solutions. We test two direct approaches and find both are harmful;
however, the pattern of failure is itself informative.

\paragraph{Approaches and results.}
On Qwen2.5-3B (GSM8K, $N{=}60$, baseline 60.0\%): High-density
prompting (system prompt instructing compact, symbolic notation)
reduces accuracy to 50.0\% ($-10$pp). Adaptive mid-step steering
(injecting a corrective phrase if LP variance deviates mid-generation)
collapses accuracy to 8.3\% ($-51.7$pp), robust across multiple prompt
formulations (Table~\ref{tab:naive}).

\begin{table}[t]
\centering\small
\caption{Na\"{i}ve density steering (Qwen2.5-3B, $N{=}60$). Adaptive
mid-step injection causes catastrophic accuracy collapse.}
\label{tab:naive}
\begin{tabular}{lcc}
\toprule
Condition & Accuracy & vs.\ Baseline \\
\midrule
Baseline (greedy CoT)        & 60.0\% & --          \\
High-density prompting       & 50.0\% & $-10.0$pp   \\
Adaptive mid-step steering   & 8.3\%  & $-51.7$pp   \\
\bottomrule
\end{tabular}
\end{table}

\paragraph{Diagnosis: density is load-bearing.}
The $-51.7$pp collapse is not caused by distraction from a badly
worded prompt. When a corrective phrase is injected mid-step, the KV
cache at that point encodes the model's accumulated computational
context, that is, the trajectory it has been following through the
reasoning chain. Injecting an inconsistent token sequence severs this
trajectory, causing incoherent continuation. Density patterns are
therefore not a stylistic overlay on reasoning; they are functionally
load-bearing, reflecting the model's internal state at each point in
generation.

This establishes a precise design principle: any viable
inference-time density intervention must \emph{preserve step-level
generation integrity}. Interventions must either act before generation
begins, or between individual tokens within a step, without
interrupting any step's completion. All methods in the following
subsections satisfy this constraint.

\subsection{Principled Nudging via Hint Injection and KV Cache
Steering}

\paragraph{Conditions.}
We evaluate four conditions on Qwen2.5-1.5B-Instruct (GSM8K,
$N{=}60$, greedy decoding):
\textbf{A (Baseline)}: standard greedy CoT.
\textbf{B (Hint)}: a NuRL-style abstract hint \citep{chen2026nurl},
generated by prompting the model to identify the problem's key
structural relationships without solving it, is prepended before the
question.\footnote{No gold labels are used at test time; the hint is
generated purely from the problem statement.}
\textbf{C (Cache Steer)}: a contrastive KV steering vector
$\mathbf{d}$ is built from the mean KV cache difference between 20 CoT
prompts and 20 direct-answer prompts at the final prompt token
position. This vector is then added to the KV cache before generation
($c_v{=}1.0$, $c_k{=}0.0$).
\textbf{D (Hint+Cache)}: combines B and C.

\begin{table}[t]
\centering\small
\caption{Principled nudging results (Qwen2.5-1.5B-Instruct,
$N{=}60$). $\Delta_{\mathrm{LP}}{=}\mathrm{Var}(C){-}\mathrm{Var}(W)$;
$\Delta_{\mathrm{Cav}}{=}\mathrm{Cav}(C){-}\mathrm{Cav}(W)$
(negative = correct solutions denser). Both signals agree in direction
across all conditions.}
\label{tab:nudging}
\begin{tabular}{lccccc}
\toprule
Cond. & Acc. & vs.\ A & $\Delta_{\mathrm{LP}}$ &
$\Delta_{\mathrm{Cav}}$ & $p$ \\
\midrule
A: Baseline     & 51.7\% & --         & $+0.002$ & $+0.006$ & n.s.  \\
B: Hint         & 46.7\% & $-5.0$pp   & $+0.017$ & $-0.008$ & 0.017 \\
C: Cache        & 53.3\% & $+1.6$pp   & $+0.005$ & $-0.027$ & n.s.  \\
D: Hint+Cache   & 65.0\% & $+13.3$pp  & $+0.004$ & $-0.007$ & n.s.  \\
\bottomrule
\end{tabular}
\end{table}

\paragraph{Results and interpretation.}
Condition D achieves $+13.3$pp, the highest accuracy of any evaluated
condition (Table~\ref{tab:nudging}). The structure of the individual
results reveals why the combination is necessary. Hint alone (B)
reduces accuracy by 5.0pp yet produces the largest LP gap ($+0.017$,
$p{=}0.017$): the hint shifts the density profile in the right
direction, but Qwen2.5-1.5B lacks the representational capacity to
capitalise on conceptual scaffolding without structural support. Cache
steering alone (C) provides structural momentum ($+1.6$pp,
$\Delta_{\mathrm{Cav}}{=}-0.027$) but lacks conceptual direction.
Their combination is effective because they address complementary
bottlenecks: the hint furnishes \emph{what} to reason about, whilst
cache steering provides the structural context in which to execute it.

LP variance and Caveman compression ratio agree in sign across all
four conditions: conditions associated with higher accuracy produce
denser correct solutions than incorrect ones. This cross-signal
agreement, between a model-internal probability signal and an
independent linguistic signal, confirms the density effect is genuine
and not a measurement artefact.

\subsection{Entropy-Gated Density Routing (EGDR)}

\paragraph{Motivation.}
The nudging pipeline above acts before generation and requires an
external hint. EGDR eliminates both requirements: it acts at the level
of individual tokens during generation, using the model's own Shannon
entropy as the only guidance signal.

At every decoding step, the model produces a probability distribution
over the vocabulary. Its Shannon entropy
$H_t = -\sum_v p_t(v) \log p_t(v)$ is a real-time, model-internal
density sensor: high $H_t$ signals a computationally dense moment
(genuine uncertainty, non-trivial decision); low $H_t$ signals
predictable filler (the model is highly confident about the next
token). This operationalisation corresponds directly to the
dual-density structure of \citet{zhao2025denser} (dense computation,
sparse explanation), derived here purely from the model's generation
process without external measurement.

\paragraph{Method.}
Let $\bar{H}$ be the mean per-token entropy over correct solutions in
Conditions A--D, and $\mathbf{d}$ the normalised KV steering vector
from Condition C. EGDR applies the following correction to the last KV
position of every transformer layer after each token is decoded:
\begin{equation}
\delta_t =
\begin{cases}
+\alpha\,\mathbf{d}      & H_t > 1.5\bar{H} \quad\text{(dense: reinforce)}\\
-0.3\alpha\,\mathbf{d}   & H_t < 0.6\bar{H} \quad\text{(sparse: moderate)}\\
0                        & \text{otherwise}
\end{cases}
\end{equation}
The asymmetric magnitudes (1.0 vs.\ 0.3) reflect the dual-density
finding: computational moments warrant stronger reinforcement than
explanatory moments warrant suppression. Thresholds
$H_{\mathrm{high}}{=}1.5\bar{H}$ and $H_{\mathrm{low}}{=}0.6\bar{H}$
are derived from Conditions A--D on the same $N{=}60$ evaluation set
(acknowledged as a limitation in the Limitations section). EGDR acts
between tokens, never mid-step, satisfying the design
principle.\footnote{\texttt{device\_map=`auto'} causes cross-device
tensor errors on multi-GPU machines; single-GPU loading via
\texttt{.to(`cuda:0')} is required. Overhead: one entropy computation
and $L$ KV additions per token, $\approx 0.3\%$ additional per-token
generation time.}

\begin{figure}[t]
\centering
\includegraphics[width=\columnwidth]{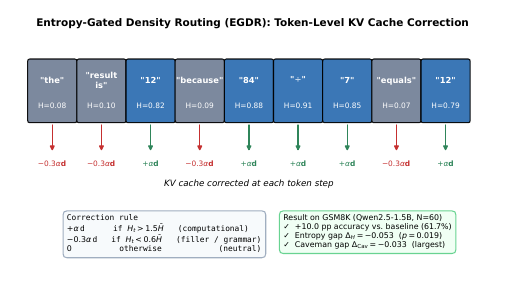}
\caption{EGDR applies a piecewise KV cache correction at every
generated token based on the model's own Shannon entropy $H_t$.
High-entropy tokens (computationally dense moments,
$H_t > 1.5\bar{H}$) receive a positive correction $+\alpha
\mathbf{d}$, reinforcing the reasoning trajectory. Low-entropy tokens
(predictable filler, $H_t < 0.6\bar{H}$) receive a small negative
correction $-0.3\alpha \mathbf{d}$. This self-supervised gating
requires no external input and never interrupts step-level generation.
At $\alpha{=}0.30$, EGDR achieves $+10.0$\,pp accuracy on GSM8K with
the most statistically significant density footprint of any evaluated
condition ($p{=}0.019$).}
\label{fig:egdr}
\end{figure}

\begin{table}[t]
\centering\small
\caption{EGDR coefficient grid search (Qwen2.5-1.5B-Instruct,
$N{=}60$). Accuracy increases monotonically; absence of degradation at
$\alpha{=}0.30$ suggests the optimum lies above the tested range.}
\label{tab:grid}
\begin{tabular}{lccc}
\toprule
$\alpha$ & Accuracy & vs.\ A & vs.\ D \\
\midrule
0.05 & 58.3\%          & $+6.6$pp  & $-6.7$pp  \\
0.10 & 55.0\%          & $+3.3$pp  & $-10.0$pp \\
0.15 & 56.7\%          & $+5.0$pp  & $-8.3$pp  \\
0.20 & 58.3\%          & $+6.6$pp  & $-6.7$pp  \\
0.30 & \textbf{61.7\%} & $+10.0$pp & $-3.3$pp  \\
\bottomrule
\end{tabular}
\end{table}

\paragraph{Novelty.}
Prior work uses per-token entropy to branch paths
\citep{li2025entropy} or select among completed traces
\citep{gwak2025revisiting}. EGDR uses entropy for continuous, in-place
KV correction during generation: self-supervised, adaptive in
magnitude, and structurally safe.

\paragraph{Results.}
Accuracy increases monotonically with $\alpha$ (Table~\ref{tab:grid}),
reaching 61.7\% ($+10.0$pp) at $\alpha{=}0.30$ with no degradation,
suggesting the optimal lies above the tested range. Hint+EGDR (G)
achieves 63.3\% ($+11.6$pp). Condition D remains the highest-accuracy
single condition (65.0\%); the gap arises because the contrastive KV
vector provides more targeted structural support than EGDR's broader
reasoning-mode direction.

Most significantly, correct solutions under EGDR exhibit substantially
lower mean per-token entropy than incorrect ones
($\Delta_H{=}-0.053$, $p{=}0.019$, Table~\ref{tab:all}), the only
result reaching $p<0.05$ on this measure across the entire experiment.
The Caveman gap ($\Delta_{\mathrm{Cav}}{=}-0.033$) is the largest of
any condition. Both signals confirm that EGDR genuinely reshapes the
model's density profile during generation, steering correct solutions
toward more focused token distributions.

\begin{table}[t]
\centering\small
\caption{Complete results across all conditions. $\Delta_H$: mean
entropy(correct)$-$mean entropy(wrong); negative = correct solutions
more confident per token. EGDR (F) is the only condition reaching
$p<0.05$ on $\Delta_H$.}
\label{tab:all}
\begin{tabular}{lccccc}
\toprule
Cond. & Acc. & vs.\ A & $\Delta_H$ & $\Delta_{\mathrm{Cav}}$ & $p$ \\
\midrule
A: Baseline   & 51.7\% & --        & --       & $+0.006$ & --    \\
B: Hint       & 46.7\% & $-5.0$pp  & --       & $-0.008$ & --    \\
C: Cache      & 53.3\% & $+1.6$pp  & --       & $-0.027$ & --    \\
D: Hint+Cache & 65.0\% & $+13.3$pp & --       & $-0.007$ & --    \\
F: EGDR       & 61.7\% & $+10.0$pp & $-0.053$ & $-0.033$ & 0.019 \\
G: Hint+EGDR  & 63.3\% & $+11.6$pp & $-0.041$ & $-0.024$ & 0.021 \\
\bottomrule
\end{tabular}
\end{table}

\subsection{Ablation Studies}

\paragraph{Density metric.}
LP variance, per-step Shannon entropy, and Caveman compression ratio
yield consistent directional patterns across all conditions,
confirming the density--correctness relationship is not an artefact of
any single operationalisation.

\paragraph{Normalisation.}
Min--max normalisation is required; z-score normalisation forces
variance to exactly 1.0 for every chain by definition, rendering
between-chain comparisons meaningless.

\paragraph{EGDR coefficient.}
Accuracy increases monotonically from $\alpha{=}0.05$ to
$\alpha{=}0.30$ (Table~\ref{tab:grid}), with a plateau at 58.3\%
between 0.05--0.20 and a jump to 61.7\% at 0.30. No degradation at the
ceiling suggests the minimum effective threshold is around
$\alpha{=}0.30$, consistent with the na\"{i}ve steering finding:
under-correction is safe but ineffective.

\section{Discussion}

\paragraph{Density as a window into model state.}
The central insight unifying all results is that information density
in CoT reasoning is not a surface stylistic property but a direct
reflection of the model's internal computational state at each
generation step. It encodes the probability mass allocation that
emerges from training, the trajectory the model is following through
its KV cache, and the degree to which each token contributes to
genuine reasoning versus grammatical scaffolding. Density cannot be
adjusted from outside without disrupting that state; it can only be
nudged through the model's own generative process or shaped before
generation begins.

\paragraph{Why EGDR succeeds where na\"{i}ve steering fails.}
Both target density, yet produce diametrically opposite outcomes. The
difference is granularity: mid-step injection disrupts the KV
trajectory at step boundaries (coarse), destroying reasoning
coherence; EGDR reinforces it token by token (fine), amplifying the
model's natural density rhythm without severing any computational
thread.

\paragraph{Paradigm determines the correct intervention direction.}
The reversal of the density--correctness relationship between SFT and
RL paradigms has a direct practical consequence: a steering method
calibrated for one regime (e.g., reducing variance for SFT models)
would be contraindicated for the other (RL models benefit from
non-uniform variance). Density-based interventions must first
establish the model's paradigm.

\paragraph{Complementarity with CoT acceleration.}
EGDR's per-token entropy signal partitions tokens into high-entropy
computational tokens (reinforced) and low-entropy grammatical tokens
(moderated), precisely the decomposition required by acceleration
methods such as SemCoT \citep{he2025semcot}, which encodes
low-information steps implicitly. Density steering and acceleration
are therefore complementary objectives applicable to the same
token-level decomposition.

\section{Limitations}

\paragraph{Scale and benchmark.}
All nudging and EGDR experiments use Qwen2.5-1.5B-Instruct and
$N{=}60$ GSM8K problems. Accuracy comparisons on $N{=}60$ have limited
statistical power; the entropy signal ($p{=}0.019$, computed over
per-token distributions) is our primary quantitative evidence.
Replication on larger sets and harder benchmarks (MATH, AIME) is
important future work.

\paragraph{Threshold circularity.}
EGDR's entropy thresholds are derived from the same $N{=}60$ set used
for evaluation; a held-out development set would be methodologically
preferable.

\paragraph{Steering vector specificity.}
The direction vector $\mathbf{d}$ encodes a broad ``reasoning mode''
(CoT vs.\ direct answer). A vector constructed from high- versus
low-entropy token contrasts within correct solutions would target
density more precisely and would likely improve both accuracy and
signal strength.

\paragraph{Cross-generational confound.}
The cross-generational comparison conflates scale with generation
(Mistral 7B is larger than GPT-2 XL); the Pythia experiment provides a
scale-controlled lower bound, but the two effects cannot be fully
isolated without matched model sizes.

\section{Conclusion}

This paper establishes that information density in LLM
chain-of-thought reasoning traces is a structural property of the
model's learned representations, traceable to an intrinsic complexity
bias confirmed by synthetic experiments (2.0--2.6$\times$ KL
asymmetry). This bias manifests as a regime-dependent correctness
signal: uniform density predicts correctness under SFT; structured
non-uniformity predicts correctness under RL distillation. Density is
load-bearing: disrupting it collapses accuracy by 51.7pp. Working with
the model's density rhythm through principled interventions, by
contrast, improves accuracy by $+13.3$pp (hint with cache steering)
and $+10.0$pp (EGDR, $p{=}0.019$). The guiding principle is that
effective use of density as a reasoning tool requires understanding it
as an expression of model state rather than as a surface stylistic
parameter.

The broader significance of this work is that it reframes information
density from a passive descriptor of reasoning quality into an active,
inference-time control surface. Because EGDR is fully self-supervised
and adds negligible overhead, it offers a practical path to improving
reasoning without retraining, additional data, or external verifiers,
and it composes naturally with existing decoding-time methods. In the
immediate term, the approach invites replication on larger evaluation
sets and harder benchmarks such as MATH and AIME, and a systematic
study of the entropy-gating thresholds on held-out data. A natural
next step is to construct steering directions that target density
explicitly rather than a broad reasoning mode, which our analysis
suggests would sharpen both accuracy and the underlying signal. Over
the longer term, density-aware inference points toward
paradigm-adaptive reasoning systems that first diagnose a model's
density regime and then apply the correspondingly beneficial
intervention, and toward a unified treatment of density-guided
steering and density-guided acceleration operating on the same
token-level structure. We believe that treating information density as
a first-class, controllable property of machine reasoning opens a
productive direction for making large language models both more
accurate and more interpretable.

\section*{Ethical Statement}

All experiments use publicly available pretrained models
(Qwen2.5-Instruct, DeepSeek R1-Distill-Qwen) and a publicly available
benchmark (GSM8K, \citealp{cobbe2021gsm8k}). No private data,
proprietary APIs, or human subjects are involved. The synthetic PCFG
corpus contains no personally identifiable information. The techniques
introduced improve reasoning accuracy without introducing capabilities
beyond those of the underlying base models; no new misuse vectors are
identified.

Authors acknowledge the assistance of
Claude\footnote{\url{https://claude.ai/}} (Opus 4.8) in drafting this
paper. Overleaf\footnote{\url{https://www.overleaf.com/}} has been
used as the \LaTeX{} editing and document-preparation environment.
Authors take ownership and responsibility for the final content.

\bibliography{references}

\newpage
\appendix

\section{Full Cross-Generational Results}

\begin{table}[h]
\centering\small
\caption{Per-model complexity preference gap $\Delta$. The Pythia
two-stage plateau (rapid growth to 410M, near-stasis at 1B--2.8B)
suggests a scaling law for complexity preference with a ceiling around
1B parameters in this architecture.}
\label{tab:crossgen-full}
\begin{tabular}{llcll}
\toprule
Model & Yr & $\Delta$ & Change & KS (next) \\
\midrule
GPT-2 Small (124M)  & 2019 & 0.3190 & --        & $p<0.01$ \\
GPT-2 Medium (355M) & 2019 & 0.3378 & $+5.9\%$  & n.s.     \\
GPT-2 Large (774M)  & 2019 & 0.3343 & $+4.8\%$  & n.s.     \\
GPT-2 XL (1.5B)     & 2019 & 0.3343 & $+4.8\%$  & --       \\
\midrule
Pythia 70M   & 2021 & 0.2935 & --         & $p<0.01$ \\
Pythia 160M  & 2021 & 0.3063 & $+4.4\%$   & $p<0.01$ \\
Pythia 410M  & 2021 & 0.3228 & $+9.9\%$   & plateau  \\
Pythia 1B    & 2021 & 0.3224 & $+9.8\%$   & n.s.     \\
Pythia 1.4B  & 2021 & 0.3321 & $+13.2\%$  & n.s.     \\
Pythia 2.8B  & 2021 & 0.3323 & $+13.2\%$  & --       \\
\midrule
GPT-2 XL        & 2019 & 0.3343 & --        & $p{=}0.187$ (\textsc{low}) \\
GPT-Neo 2.7B    & 2021 & 0.3816 & $+14.2\%$ & $p{=}1.3\times10^{-47}$    \\
Mistral 7B v0.3 & 2023 & 0.4054 & $+21.3\%$ & --                         \\
\bottomrule
\end{tabular}
\end{table}

\end{document}